\documentclass[11pt,a4paper]{article}

\usepackage[affil-it]{authblk}
\usepackage[margin=3.0cm]{geometry}
\usepackage{amsmath,amssymb,amsfonts,upref}
\usepackage{amsthm}
\usepackage{mathtools}
\usepackage{tabularx}
\usepackage{soul}

\usepackage{listings}
\usepackage{xcolor}
\title{Philosophy-Guided Modelling and Implementation of Adaptation and Control in Complex Systems}
\author[1]{Olivier Del Fabbro}
\affil[1]{ETH Zurich, Chair for Philosophy, Clausiusstrasse 49, 8092 Zurich, Switzerland}
\author[2]{Patrik Christen\footnote{Corresponding author: patrik.christen@fhnw.ch. The authors contributed equally to this work.}}
\affil[2]{FHNW University of Applied Sciences and Arts Northwestern Switzerland, Institute for Information Systems, Riggenbachstrasse 16, 4600 Olten, Switzerland}

\date{31 August 2020 (last revised 25 September 2021)}

\begin{document}

\maketitle

\abstract{Control was from its very beginning an important concept in cybernetics. Later on, with the works of W. Ross Ashby, for example, biological concepts such as adaptation were interpreted in the light of cybernetic systems theory. Adaptation is the process by which a system is capable of regulating or controlling itself in order to adapt to changes of its inner and outer environment maintaining a homeostatic state. In earlier works we have developed a system metamodel that on the one hand refers to cybernetic concepts such as structure, operation, and system, and on the other to the philosophy of individuation of Gilbert Simondon. The result is the so-called allagmatic method that is capable of creating concrete models of systems such as artificial neural networks and cellular automata starting from abstract building blocks. In this paper, we add to our already existing method the cybernetic concepts of control and especially adaptation. In regard to the system metamodel, we rely again on philosophical theories, this time the philosophy of organism of Alfred N. Whitehead. We show how these new meta-theoretical concepts are described formally and how they are implemented in program code. We also show what role they play in simple experiments. We conclude that philosophical abstract concepts help to better understand the process of creating computer models and their control and adaptation. In the outlook we discuss how the allagmatic method needs to be extended in order to cover the field of complex systems and Norbert Wiener's ideas on control.}


\newpage
\section{Introduction}

Cybernetics was from its very beginning a scientific endeavor that tried to bring together as many disciplines as possible. Already in the 1943 written paper, \emph{Behavior, Purpose and Teleology}, Norbert Wiener, Arturo Rosenblueth, and Julian Bigelow \cite{Rosenblueth.1943}, mixed concepts from physiology, neuropsychology, engineering, and even philosophy (teleology) in order to highlight analogies between living beings and machines. Active and purposeful behaviour leads to an output of signals which can be fed back to the system so that the teleology, i.e. the end, aim, goal, finality, of the system under consideration is altered. In the end, it does not matter if that system is the human nervous system or an anti-aircraft defence system.

Yet, what seems to be missing in that publication and also in Wiener’s 1948 published monograph, \emph{Cybernetics} \cite{Wiener.1948}, is the interpretation of feedback and control in biological terms. It was left to others in the cybernetic community such as W. Ross Ashby to introduce more systematically biological concepts such as adaptation, environment, organism, survival, and struggle for existence \cite{Ashby.1952}. For Ashby, adaptation is a form of behaviour that is capable of maintaining essential variables within certain physiological limits in order to establish homeostatic stability \cite[p.~58]{Ashby.1952}. Wiener, however, anticipated such interpretation by elucidating the concept of homeostasis in his own view: adaptation is then a process of internal regulation of a system through control and feedback mechanisms in order to adjust to the influences of the environment and to establish a homeostatic state \cite[pp.~114--115]{Wiener.1948}. In cybernetics, the concept of adaptation is thus directly bound to the concept of control and feedback and the establishment of a homeostatic state.

But, what counts for biology can also be said for other fields. The famous Macy Conferences show perfectly well how sociology, psychoanalysis, information theory, etc. could adhere to the cybernetic project \cite{Pias.2003}. Even humanities got influenced by the cybernetic movement.\footnote{Please see \cite{Christen.2019} for more detailed information.} One of these philosophers influenced by cybernetics was the Frenchman Gilbert Simondon (1924--1989) \cite{DelFabbro.2021,Guchet.2010}. Simondon used concepts from cybernetics such as system, structure, operation, feedback, information, and so on in order to establish a metaphysics of individuation that should be able to describe all kinds of phenomena and objects: wave-particle dualism in physics, chemical growth of crystals, biological beings such as polyps, psychological behaviour such as emotivity, social emergence of groups and technical objects such as combustion engines and vacuum tubes \cite{Simondon.2020}. Yet, while Simondon only worked with pen and paper to write down his metaphysics in the 1950’s, technological advancements in computer modelling allowed the authors of this paper to implement Simondon’s abstract concepts into a system metamodel \cite{Christen.2019}. Abstract concepts such as structure and operation helped to define generic building blocks of a system metamodel such as entities and update function. From these abstract building blocks, it is then possible to construct concrete models such as cellular automata and artificial neural networks. The abstract building blocks are described in a metamodel of a system and are part of the so-called allagmatic method. Its name is in reference to Simondon and means change or transition \cite[pp.~11--16]{Chateau.2008}. It proceeds through different so-called regimes or stages: virtual, metastable, actual, or in other words, from abstract generic concepts to concrete models.

What up to this point, however, was missing in this system metamodel is the concept of adaptation as it was introduced by Wiener and Ashby. In section 2 we give a short summary of the already developed system metamodel. In section 3 we introduce the concept of adaptation, i.e. adaptation function. Adaptation means here a whole process in which the behaviour of a system is adjusted due to the evaluation of a target or aim that should be reached.

However, by adding the concept of adaptation to the system metamodel, further meta-theoretical expansion by philosophical concepts is required. Here, we rely on the concepts of adaptation by the mathematician and philosopher Alfred N. Whitehead, whose philosophy of organism is close to Simondon’s philosophy of individuation \cite{Debaise.2017}. Especially, for future work on the intricacy and complexity of systems -- as we will show in section 8 -- Whitehead’s meta-theoretical work will be of great help. At the same time, we show that Whitehead’s concepts of adaptation are compatible with cybernetic vocabulary \cite{Hampe.1990}. We introduce Whitehead’s concepts in section 4 of this paper. In section 5, we then show how Whitehead’s concepts find their place within the allagmatic method. In section 6 of the paper, we give an account of adaptation in simple experiments with cellular automata and artificial neural networks. The experiments do not only show that it is possible to create concrete models such as cellular automata and artificial neural networks from abstract building blocks, the introduction of philosophical meta-theoretical concepts also allow to give guidance for interpretation through the different regimes. Once operating, the concepts have a specific well-defined place within the allagmatic method. We show this in section 6 and in the discussion 7. Section 6 also describes how control mechanisms are present within the allagmatic method and how they relate to the adaptation function. Control is the manner of how the update function of a system is being steered.

Even though simple, the experiments show one important aspect in today’s use of computer models. Once the models attain a target they do not continue to evolve \cite{Stanley.2019}. For Wiener, however, and Simondon and Whitehead alike, control in real systems is always an assembly of controllers and governors that collectively try to maintain homeostasis: body temperature, waste products in the digestive system, chemical defences against infection, heart rate, blood pressure and so on \cite[pp.~114--115]{Wiener.1948}. There is not a single control mechanism. In the outlook 8, we show how the meta-theoretical concepts of Whitehead are capable of giving guidance to implement future models with multiple control mechanisms.

\section{The Allagmatic Method}

Using the recently developed and implemented system metamodel for modelling systems, we directly \cite{Christen.2019} and automatically \cite{Christen.2020} created cellular automata and artificial neural networks with the abstract and generic model building blocks of the system metamodel of the allagmatic method. The system metamodel is inspired by the philosophical concepts structure and operation as proposed by Simondon to describe a system \cite{Simondon.2013}. Each system, natural or artificial, consists of at least one structure capturing the spatial dimension and at least one operation capturing the temporal dimension. We formally defined a model of a system $\mathcal{SM}$ at the most abstract level with the tuple $\mathcal{SM} \coloneqq (\hat{s}_1,\hat{s}_2,\hat{s}_3,\dots,\hat{s}_s,\hat{o}_1,\hat{o}_2,\hat{o}_3,\dots,\hat{o}_o)$, where $\hat{s}_i$ are structures of set $S$ and $\hat{o}_j$ are operations of set $O$, thus $\hat{s}_i\in S \land \hat{o}_j\in O$ \cite{Christen.arXiv.2020}.

More concretely, structure $S$ and operation $O$ are described in more detail by further dividing them into more specific but still general concepts. Guided by the systems view, systems are made up by interacting entities defined with an entity $e$-tuple $\mathcal{E}=(\hat{e}_1,\hat{e}_2,\hat{e}_3,\dots,\hat{e}_{e})$, where $\hat{e}_i\in Q$ with $Q$ being the set of $k$ possible states \cite{Christen.arXiv.2020}. These entities are interacting with each other within a certain neighbourhood or milieu defined with the milieus $e$-tuple $\mathcal{M}=(\hat{\mathcal{M}}_1,\hat{\mathcal{M}}_2,\hat{\mathcal{M}}_3,\dots,\hat{\mathcal{M}}_{e})$, where $\hat{\mathcal{M}}_{i}=(\hat{m}_1,\hat{m}_2,\hat{m}_3,\dots,\hat{m}_m)$ is the milieu of the $i$-th entity $\hat{e}_i$ of $\mathcal{E}$ consisting of $m$ neighbours of $\hat{e}_i$ \cite{Christen.arXiv.2020}. Additionally, there can be structures for storing some information used by operations. The most basic operation updates the states of an entity depending of the states of neighbouring entities following some local rules. We defined it as an update function $\phi : Q^{m+1} \rightarrow Q$ and introduced the structure $\mathcal{U}$ storing the rules or logic of the function \cite{Christen.arXiv.2020}.

\begin{figure}[htbp]
	\centerline{\includegraphics[width=1.0\textwidth]{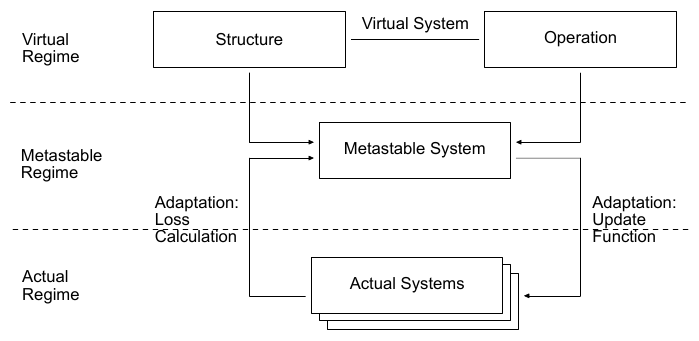}}
	\caption{The allagmatic method consisting of a system metamodel that is abstractly described in the virtual regime, concretised with parameters in the metastable system, and run or executed in the actual regime. It starts with the object creation and thus the transition from the virtual to the metastable regime, continues through the metastable and actual regimes, and ends with the transition from the actual back to the metastable regime. Adaption is occurring between the metastable and actual regimes.}
	\label{fig1}
\end{figure}

The system metamodel is applied in the allagmatic method \cite{Christen.2019} following three different regimes (Fig.~\ref{fig1}): In the virtual regime, the system metamodel is described abstractly. In the metastable regime, structure and operation are combined and fed with model parameters. This creates an actual system that can be run in the actual regime.

\section{Formalism and Implementation of Adaptation}

From a formal point of view, we introduce here a further operation acting on the structure $\mathcal{U}$ capable of adapting or changing it. We defined it as an adaptation function $\psi$ with its own logic stored in the structure $\mathcal{A}$ and adaptation end or target stored in $\mathcal{P}$ \cite{Christen.arXiv.2020}. Structure is therefore extended with adaptation and defined by $S \coloneqq \{\mathcal{E}, Q, \mathcal{M}, \mathcal{U}, \mathcal{A}, \mathcal{P}, \dots, \tilde{s}_s \}$, where $\tilde{s}_i$ are possible further structures, and operation by $O \coloneqq \{ \phi(\hat{e}_i,\hat{\mathcal{M}}_i,t,\mathcal{U}),\psi(g,\mathcal{A},\mathcal{P},l),\dots, \tilde{o}_o \}$, where $t$ denotes the number of time steps and $\bar{t}$ the current time step, $g$ the number of adaptation iterations and $\bar{g}$ the current adaptation iteration, and $l$ the loss tolerance \cite{Christen.arXiv.2020}. The model of a system $\mathcal{SM}$ is finally defined more precisely and formally capturing adaptation with $\mathcal{SM} \coloneqq (\mathcal{E},Q,\mathcal{M},\mathcal{U},\mathcal{A},\mathcal{P},\dots,\hat{s}_s,\phi,\psi,\dots,\hat{o}_o)$, where $\hat{s}_i\in S \land \hat{o}_j\in O$ \cite{Christen.arXiv.2020}.

In this study, we added the adaptation operation to the system metamodel implementation as well as reimplemented the rest of the system metamodel in accordance with the definitions above and described in detail in another study that focused on the formalism of the system metamodel \cite{Christen.arXiv.2020}. As in the previous implementation \cite{Christen.2019}, guidance from the philosophy of Simondon \cite{Simondon.2013}, object-oriented programming and template meta-programming in C++ was used \cite{Stroustrup.2013,Alexandrescu.2001}. Classes are used to implement structures as data member and operations as member functions of classes defined in the virtual regime. Since the model parameters become concrete only through parametrisation in the metastable regime, generic types are used in the virtual regime implemented with the C++ vector template. Once an object is created and all parameters are initialised, it is run in the actual regime. The object creation starts with the transition from the virtual to the metastable regime, continuous through the metastable and actual regimes, and ends with the transition from the actual back to the metastable regime. Adaption is occurring between the metastable and actual regimes (Fig.~\ref{fig1}).

\section{Meta-Theoretic Description of Adaptation}

Before describing how the adaptation function is integrated into the allagmatic method and system metamodel, and how it operates in concrete models such as cellular automata and artificial neural networks, it is necessary to give philosophical descriptions of adaptation on a meta-theoretical level. These meta-theoretical descriptions of adaptation in turn follow the philosophy of organism of Alfred N. Whitehead, because his philosophy is on the one hand similar to that of Simondon \cite{Debaise.2017}, and on the other hand Whitehead has given a more thorough analysis and deployment of the concept of adaptation.

Already in our previous studies \cite{Christen.2019,Christen.2020,Christen.arXiv.2020}, the concept of entity, which constitutes the very basic structure of any system, is borrowed from Whitehead’s philosophy of organism, when he uses the concept of actual entity to denote “the final real things of which the world is made up.” \cite[p.~18]{Whitehead.1929} Similar to our system metamodel in which entities form together a system, Whitehead believes that actual entities form together a so-called nexus. 

The important point in regard to adaptation is that it gives the entities a target, goal, or end (a telos in philosophical terms) that needs to be attained (or a problem to be solved). Hence, when the behaviour of entities is structured through an adaptive operation (an adaptation function) and the given end of that function is attained, a so-called society is formed. Society, as it is defined by Whitehead, should therefore not be understood in the narrow and common sense of a social gathering of humans studied in sociology. In this sense, Whitehead writes: “The most general examples of such societies are the regular trains of waves, individual electrons, protons, individual molecules, societies of molecules such as inorganic bodies, living cells, and societies of cells such as vegetable and animal bodies.” \cite[p.~98]{Whitehead.1929} Both, the forming of a society and of a nexus are processes of emergence, but it is the presence of an aim which makes the difference between a nexus and a society. Today, in agent-based modelling such a society is called a system whereas in swarm intelligence it is called a swarm and in artificial neural networks it represents the network. Lastly, what Whitehead calls society is termed by Simondon an individuation \cite{Debaise.2017}. We will use Whitehead’s terminology of society and nexus.

For both, a nexus and a society, entities interact with each other in order to follow a certain path of behaviour (update function) but only in a society an end is attained (adaptation function). As shown above in the quote, this phenomenon can, according to Whitehead, also be observed in biology and physics, when molecules form a cell and waves and electrons an electromagnetic field, respectively. Nex\={u}s (the plural of nexus) are thus defined as actual systems that did not reach the targeted end. When a society has attained a certain end, “it is self-sustaining.” \cite[p.~89]{Whitehead.1929}

\section{Adaptation and Control in the Allagmatic Method}

\subsection{Update Function}

Here, we suggest to integrate adaptation into the system metamodel and the allagmatic method with the guidance of Whitehead’s metaphysics. However, we first explain the details of the update function that was created and implemented according to the philosophy of Simondon \cite{Simondon.2013} in earlier studies \cite{Christen.2019,Christen.2020,Christen.arXiv.2020}.

The distinction of structure and operation in a system is borrowed from Simondon. These operations act upon structures in three different regimes (Fig.~\ref{fig1}), which provides a general framework for describing the emergence of a system in terms of structure and operation in the allagmatic method. At the most abstract level in the virtual regime, the update function is an operation of the system model $\mathcal{SM}$ and is defined as a mathematical function $\phi(\hat{e}_i,\hat{\mathcal{M}}_i,t,\mathcal{U}) : Q^{m+1} \rightarrow Q$, where all its parameters remain abstract in that regime. Therefore, $\phi$ is implemented as a member function of the class \texttt{SystemModel} formally referred to as $\mathcal{SM}$. The function itself is specific to the application and is thus not defined abstractly. Several existing and concrete update functions as used in elementary cellular automata \cite{Wolfram.2002} and simple feedforward artificial neural networks \cite{Russel.2010} are thus implemented in the class \texttt{SystemModel}. The process starts with the creation of a \texttt{SystemModel} object with a specific type defining $Q$ and thus the transition between the virtual and metastable regime. Please also note that the parameters are the same for each of these functions and are implemented as generic types and with dynamic sizes to achieve the abstract description in the virtual regime. In the metastable regime, the main program further instantiates the \texttt{SystemModel} object, where the types are defined and therefore concretised through template meta-programming. A C++ vector with objects of the class \texttt{Entity} is created next and assigned to the data member \texttt{entities} of the class \texttt{SystemModel}. With that, the system model is further concretised with the number of entities. In the same way, $\mathcal{M}$ and $\mathcal{U}$ are created and assigned to the data member \texttt{milieus} and \texttt{updateRules} of the \texttt{SystemModel} object, respectively. With this creation, the system is further concretised with the specific neighbourhood and thus topology of the entities as well as the specific update rule. By providing the number of time steps $t$, the creation of the \texttt{SystemModel} object in the metastable regime is finished. In the actual regime, the \texttt{SystemModel} object is then run or executed with the specified parameters for $t$ steps.

\subsection{Adaptation Function}

To run computations that are capable of adapting, we extended our system metamodel with a further model building block called adaptation function and formally referred to as $\psi$ in the present study guided by the metaphysics of Whitehead \cite{Whitehead.1929}. It is now described in a similar way as the update function in the previous paragraph in the context of the allagmatic method and its implementation in the system metamodel as well as the guidance from Simondon’s and Whitehead’s metaphysics with respect to adaptation. Again we start with the distinction of structure and operation acting in the three different regimes. The following description of the different steps starting from the virtual regime leading to the creation of concrete models in the actual regime shows how the concept of adaptation is itself created and at the same time operating within the allagmatic method.

At the most abstract level in the virtual regime, the adaptation function is an operation of the system model $\mathcal{SM}$ and is defined as a mathematical function $\psi(g,\mathcal{A},\mathcal{P},l)$, where all its parameters remain abstract in that regime. Therefore, $\psi$ is implemented as a member function of the class \texttt{SystemModel} formally referred to as $\mathcal{SM}$. The function itself is specific to the application and update function and is thus not defined abstractly. We implemented specific adaptation functions for cellular automata \cite{Wolfram.2002} and simple feedforward artificial neural networks \cite{Russel.2010} in the class \texttt{SystemModel}. E.g. for elementary cellular automata, there are $2^8$ possible rules in the update function. We thus implemented an adaptation function that randomly selects one of these possible rules without considering any error or fitness measure. In contrast, for the update function of artificial neural networks, an adaptation function that adapts the weights of the network taking into account the error according to the so-called perceptron learning rule \cite{Russel.2010}. Both adaptation functions are implemented in the class \texttt{SystemModel}. Please note that although some parameters might not be used in specific functions, the ones that are used are the same for each of these functions and are implemented as generic types and with dynamic sizes to achieve the abstract description in the virtual regime. Transitioning from the virtual to the metastable regime (Fig.~\ref{fig1}), identical to the update function description above, the main program instantiates a \texttt{SystemModel} object, where the types are defined and therefore concretised through template meta-programming. A C++ vector with objects of the class \texttt{Entity} is created next and assigned to the data member \texttt{entities} of the class \texttt{SystemModel}. With that, the system is further concretised with the number of entities. In the same way, $\mathcal{M}$ and $\mathcal{U}$ are created and assigned to the data member \texttt{milieus} and \texttt{updateRules} of the \texttt{SystemModel} object, respectively. With this creation, the system is further concretised with the specific neighbourhood and thus topology of the entities as well as the specific update rule. By providing the number of time steps $t$, the creation of the \texttt{SystemModel} object in the metastable regime is finished if no adaptation is considered.

To include adaptation, the number of adaptation iterations $g$, the structures for adaptation rules $\mathcal{A}$ and adaptation end $\mathcal{P}$, and loss tolerance $l$ are defined and with that the system further concretised. Please note that $\mathcal{A}$ as well as $\mathcal{U}$ can be explicitly specified with the given structures or directly be implemented with the respective functions. In the actual regime (Fig.~\ref{fig1}), the \texttt{SystemModel} object is run or executed with the specified parameters, that is with concrete number of entities $e$, milieu topology $\mathcal{M}$, update function $\phi$, number of time steps $t$, adaptation function $\psi$, and number of adaptation iterations $g$. Once $t$ is achieved, adaptation takes place and thus one computation process ends (Fig.~\ref{fig1}).

\subsection{Adaptation and Control}

According to Whitehead, at this stage a nexus has been formed if the end $\mathcal{P}$ is not yet reached or a society if the end $\mathcal{P}$ is reached. System states are stored as member data \texttt{systemStates} of the \texttt{SystemModel class}. To decide whether or not the end has been reached, first and in any case of the adaption process, that is for every adaptation function $\psi$ specified, a loss $l$ is calculated. Loss relates to different terms such as error and fitness. It is application specific and needs to be specified as such for the different adaptation functions $\psi$. In our computer experiments, we are searching for a specific sequence called target or end captured by the structure $\mathcal{P}$ and consisting of zeros and ones. The loss $l$ is in this case calculated by the mean square error when comparing each position of the resulting sequence after $t$ steps with the target $\mathcal{P}$. Second and again in any case of the adaptation process, the update function $\phi$ is adapted. \emph{This step is what can be regarded as control in our allagmatic method}. Specifically, the rules $\mathcal{U}$ according to which the update function $\phi$ is working are updated. With that, the relation between entities is adapted since the entities will update according to different rules based on the entities’ states in the milieu. Thus, an entity’s relation to its milieu is at every adaptation iteration different. Additionally, but not in any case of specific adaptation functions, the adaptation of the update function $\phi$ takes into account an error, fitness, or learning $l$. In our computer experiments, the adaptation function for elementary cellular automata does not take into account any error or learning, it randomly selects from a given pool of update rules. In contrast, the adaptation function of the artificial neural network, takes into account the error of each neuron and adapts the network weights according to a learning rule. Once adaptation is finished, a new process starts where an actual system with the adapted update function is created in the metastable system, which is then run or executed again in the actual regime until the number of adaptation iterations $g$ or a certain loss $l$ is reached.

\emph{Control} is thus part of the whole adaptation process, i.e. when within the operation of adaptation function $\psi$, control can be exercised on the update function $\phi$ of a system. In other words: the behaviour of an entity, or nexus of entities is being influenced and manipulated.

\section{Adaptation and Control in Cellular Automata and Artificial Neural Networks}

\subsection{Simple Experiments}

Two simple experiments were performed with the newly implemented system metamodel of the allagmatic method. In the first experiment $\mathcal{SM}^{CA}$, a cellular automaton was created and concretised with 31 entities $e$, Boolean entity states $Q=\{0,1\}$, a milieu $\mathcal{M}$ consisting of $m=2$ neighbouring entities, 15 time steps $t$, an update function $\phi : Q^{m+1} \rightarrow Q$, initial conditions $\mathcal{E}^{0}=0000000000000001000000000000000$, an adaptation function $\psi$ randomly selecting a rule, 100000 adaptation iterations $g$, a target or end $\mathcal{P}=(1101011001111101000000000000000)$ representing Wolfram's rule 110 \cite{Wolfram.2002} at time step 15, a loss calculation according to the mean square error when comparing each entity of the current system at $t=15$ with the target $\mathcal{P}$, and a loss $l=0.001$. In the second experiment $\mathcal{SM}^{ANN}$, a simple feedforward artificial neural network \cite{Russel.2010} was created and concretised with 31 neurons per layer and 15 layers. The other parameters were concretised in the same way as in the cellular automaton experiment with the exception of the update and adaptation functions. A weighted input function and threshold action function were implemented in the update function and with the perceptron learning rule, a more advanced adaptation function was implemented to account for learning \cite{Russel.2010}.

\subsection{Formalism and Program Code}

The concretisation process can now be observed and followed starting with the transition from the virtual to the metastable regime (Fig.~\ref{fig1}). It is in the following described by providing a description and formalism of the system metamodel followed by its implementation in C++ program code.

\subsubsection{Virtual regime}\hfill \break
\indent a) The class \texttt{Entity} is given with no concrete structures and operations. It has a generic data member \texttt{state} representing the state of an entity $\hat{e}_i$ of $\mathcal{E}$, where $\hat{e}_i\in Q$. In the virtual regime, this class is stored in the program code and is independent of the concrete system modelled.
		
\begin{lstlisting}
template <typename Q>
class Entity
{
	Q state;
};
\end{lstlisting}
		
b) The class \texttt{SystemModel} is given with no concrete structures and operations. It has generic data members \texttt{systemStates}, \texttt{milieus}, \texttt{updateRules}, \texttt{adaptationRules}, and \texttt{targetStates} denoted $\mathcal{E}^{(0,\dots,t)}$, $\mathcal{M}$, $\mathcal{U}$, $\mathcal{A}$, $\mathcal{P}$, respectively, in the system metamodel. It also has member function implementations of update functions \texttt{updateFunction()} denoted $\phi$ and adaptation functions \texttt{adaptationFunction()} denoted $\psi$ that are concrete in their operation but generic in terms of data types. In the virtual regime, this class is stored in the program code and is independent of the concrete system modelled although the functions contain concrete operations. These functions are selected during concretisation in the metastable regime.
		
\begin{lstlisting}
template <typename Q>
class SystemModel
{
	std::vector<std::vector<Entity<Q>>> systemStates;
	std::vector<std::vector<int>> milieus;
	std::vector<Q> updateRules;
	std::vector<Entity<Q>> targetStates;
	Q updateFunctionCA(...) {...}
	std::vector<Q> adaptationFunctionCA() {...}
	Q updateFunctionANN(...) {...}
	std::vector<Q> adaptationFunctionANN() {...}
};
\end{lstlisting}
		
\subsubsection{Metastable regime}\hfill \break
\indent a) An object of the class \texttt{SystemModel}, denoted by $\mathcal{SM}$ in the system metamodel, is created giving a concrete data type of the data member \texttt{state}, represented by the set $Q$ of $k$ possible states. At this point the transition from the virtual to the metastable regime happens and therefore also the concretisation process begins.

Experiments: $\mathcal{SM}^{CA}$ and $\mathcal{SM}^{ANN}$ are created with $Q=\{0,1\}$.

\begin{lstlisting}
SystemModel<bool> systemModelCA;
SystemModel<bool> systemModelANN;
\end{lstlisting}

b) Objects of the class \texttt{Entity} are created. They might be created in an array-like data structure represented by the $e$-tuple $\mathcal{E}$, where the number of entities $e$ is given. Initial conditions according to the concrete values of $Q$, $k$, and $e$ are created in $\mathcal{E}^{(0)}$ and stored to data member \texttt{systemState} of the \texttt{SystemModel} object denoted $\mathcal{SM}$.
		
Experiments: $\mathcal{E}^{(0)} = (0000000000000001000000000000000)$ is created with $e=31$ and $\hat{e}_i\in Q$, and added to $\mathcal{SM}^{CA}$ and $\mathcal{SM}^{ANN}$.
		
\begin{lstlisting}
std::vector<Entity<bool>> initialEntities(numberOfEntities);
...
systemModelCA.setSystemEntities(initialEntities);
systemModelANN.setSystemEntities(initialEntities);
\end{lstlisting}

c) Milieus $\mathcal{M}$ are created and stored to data member \texttt{milieus} of the \texttt{SystemModel} object denoted $\mathcal{SM}$.

Experiments: $\mathcal{M}=(\hat{\mathcal{M}}_1,\hat{\mathcal{M}}_2,\hat{\mathcal{M}}_3,\dots,\hat{\mathcal{M}}_{e})$ is created and added to $\mathcal{SM}^{CA}$ and $\mathcal{SM}^{ANN}$. $\hat{\mathcal{M}}_{i}=(\hat{m}_1,\hat{m}_2)$, where $\hat{m}_1$ is the neighbour to the left and $\hat{m}_2$ to the right of $\hat{e}_i$, and $\hat{m}_1,\hat{m}_2\in\mathbb{N}$.

\begin{lstlisting}
std::vector<std::vector<int>> milieus(numberOfEntities);
...
systemModelCA.setSystemMilieus(milieus);
systemModelANN.setSystemMilieus(milieus);
\end{lstlisting}

d) Update rules $\mathcal{U}$ and update function $\phi$ are created according to the concrete values of $Q$, $k$, and $\mathcal{M}$. $\mathcal{U}$ is stored to data member \texttt{updateRules} of the \texttt{SystemModel} object denoted $\mathcal{SM}$. $\phi$ is selected from concretely implemented member functions \texttt{updateFunction()} of the \texttt{SystemModel} object. Notice that these functions are implemented with template programming to account for different data types and milieus.

Experiments: $\mathcal{U}^{CA} = (01101110)$ is created and stored in $\mathcal{SM}^{CA}$. $\phi^{CA}$ for elementary cellular automata \cite{Wolfram.2002} is selected in $\mathcal{SM}^{CA}$. $\phi^{ANN} = \alpha\circ \beta{_j}$, where $\alpha = \begin{cases}
0 \text{ if } 0.5 > \beta_j\\
1 \text{ if } \beta_j \geq 0.5
\end{cases}$ and $\beta{_j}=\sum_{i=1}^m{\omega_{i,j} a_i} + \omega_b$ \cite{Russel.2010} are selected in $\mathcal{SM}^{ANN}$. $m=3$ is the milieu or number of incoming activation signals, $\omega$ are randomly initialised weights, $a_i$ are incoming activation signals, and $\omega_b$ is a randomly set bias weight.

\begin{lstlisting}
int numberOfRulesCA = 8;
std::vector<bool> updateRulesCA(numberOfRulesCA);
...
systemModelCA.setSystemRules(updateRulesCA);
Q updateFunctionCA(...) {...}
Q updateFunctionANN(...) {...}
\end{lstlisting}

e) Optionally, adaptation rules $\mathcal{A}$, target $\mathcal{P}$, and adaptation function $\psi$ are created according to the concrete values of $\mathcal{U}$, $e$, $Q$, and $k$. $\mathcal{A}$ and $\mathcal{P}$ are stored to data member \texttt{adaptationRules} and \texttt{targetStates} of the \texttt{SystemModel} object denoted $\mathcal{SM}$, respectively. $\psi$ is selected from concretely implemented member functions \texttt{adaptationFunction()} of the \texttt{SystemModel} object. Notice that these functions are implemented with template programming to account for different data types and milieus.
		
Experiments: $\mathcal{P}=(1101011001111101000000000000000)$ is created and stored in $\mathcal{SM}^{CA}$ and $\mathcal{SM}^{ANN}$. $\psi^{CA}$ is selected in $\mathcal{SM}^{CA}$, where $\mathcal{U}$ is adapted in a random manner. It is thus randomly controlled. $\psi^{ANN} = \omega_{i,j} \leftarrow \omega_{i,j} + r(y-a_j)a_i$ is selected in $\mathcal{SM}^{ANN}$, where $r=0.01$ is the learning rate, $y$ the target or desired activation signal, $a_j$ the outgoing activation, and $a_i$ the incoming activation \cite{Russel.2010}. Here, control is not random, instead it accounts for a certain target and learning strategy. Loss is calculated by the mean square error comparing each element of $\mathcal{P}$ with $\mathcal{E}^{(t)}$.

\begin{lstlisting}
std::vector<Entity<bool>> targetStates(numberOfEntities);
...
systemModelCA.setSystemTarget(targetStates);
systemModelANN.setSystemTarget(targetStates);
...
std::vector<Q> adaptationFunctionCA() {...}
void adaptationFunctionANN(learningRate) {...}
\end{lstlisting}

\subsubsection{Actual regime}\hfill \break
\indent a) An object of the class \texttt{SystemModel}, denoted by $\mathcal{SM}$ in the system metamodel, is run in a simulation of $t$ discrete time steps.
		
Experiments: $\mathcal{SM}^{CA}$ and $\mathcal{SM}^{ANN}$ are run for $t=15$ iterations.
		
\begin{lstlisting}
int numberOfIterations = 15;
void runSimulationCA(numberOfIterations) {...}
void runSimulationANN(numberOfIterations) {...}
\end{lstlisting}		
		
b) An object of the class \texttt{SystemModel}, denoted by $\mathcal{SM}$ in the system metamodel, is run in an adaptation of $g$ iterations.
		
Experiments: $\mathcal{SM}^{CA}$ and $\mathcal{SM}^{ANN}$ are adapted for $g=100000$ iterations. Notice that between each of the adaptation iterations, $t$ update iterations are simulated. Loss is calculated in each adaptation iteration and the program stops if either the current adaptation iteration $\bar{g}$ reaches the number of adaptation iterations $g$, thus $\bar{g}=g$, or the calculated loss is less than $l=0.001$.
		
\begin{lstlisting}
int numberOfAdaptations = 100000;
void runAdaptationCA(numberOfAdaptations) {...}
void runAdaptationANN(numberOfAdaptations) {...}
\end{lstlisting}

\subsection{Adaptation and Philosophical Interpretation}

The simple experiments show that adaptation is not a static concept, but a process occurring between the metastable and actual regime and therefore it is a process that happens within the framework of the allagmatic method. Only once the end is attained a society occurs but the different adaptation processes create continuously new societies and nex\={u}s. Since the creation of a model of a system, that is $\mathcal{SM}$ or a \texttt{SystemModel} object, is implemented in such a way that it represents societies as described by Simondon and Whitehead, it can be followed meticulously. This is achieved by object-oriented programming and template meta-programming of the system metamodel in the allagmatic method.

Section 5 shows how the different concepts are created and related one to another. It combines formal-mathematical and conceptual-philosophical descriptions with implementation into program code. At no time, the concept of adaptation lies outside of any sort of description. Its behaviour can always be linked to a theoretical framework. This also counts for the processual and operative aspect of program code and thus the emergence of adaptation as a concept.

Notice that the emphasis of the allagmatic method does not lie solely on the computed result, i.e. the society. At the end of all computations in order to find a society that reaches the given target, the created realm of systems consists of multiple nex\={u}s and one society. These nex\={u}s and society in turn are not themselves forming yet another nexus or society because they exist independently from one another. There is no relation between the entities of one nexus to that of another of any kind. There is thus a spatial demarcation. Notice also that the implementation did not allow the nex\={u}s and society to further develop -- even though it would be computationally possible. The created societies and nex\={u}s do not further develop, change, or adapt further. They still exist but are completely silent in behaviour. Notice furthermore that inside-outside distinctions are possible to make. Every single nexus and society has an interior with a certain number of entities and some of these entities forming the border of the nexus.

\section{Discussion}

We have shown how an adaptation function can be introduced into our allagmatic method and how a difference can be made between two types of operations: update function and adaptation function. Adaptation as we understand it, was defined meta-theoretically with the help of concepts by the philosophy of organism of Whitehead. Furthermore, these philosophical definitions are compatible with definitions of adaptation and control by Wiener and Ashby. Adaptation is the perpetual change of a system's update rules in order to attain a certain end or goal. In this sense, adaptation is linked to control and we have shown that control is executed in our allagmatic method within the adaptation function operating on the update function.

The additional value of the introduction of adaptation as adaptation function in our allagmatic method is important because it gives a processual description of adaptation. This is important because it shows that adaptation is not outside of the system. It is an integrative feature of systems themselves. Adaptation is thus not an obscure concept, but a concept that is part of a whole system, playing a specific role in the creation of societies and nex\={u}s. The allagmatic method is thus in line with the goal of Whitehead's metaphysics that provides generic notions which add lucidity and clarity to the apprehension of (in this case) computer models. 

All in all, the allagmatic method and its description of adaptation is compatible with standard methods in evolutionary computation \cite{Holland.1992,Baeck.2000,Rechenberg.2000}, artificial intelligence \cite{Russel.2010}, and complex systems research \cite{Thurner.2018,Holland.2012,Mitchell.2009,Kauffman.1993} in general. Adaptation operates on the update function and calculates a loss, error, or fitness. The implementation of the concept of adaptation is thus not a specificity of the allagmatic method. Its specificity in the allagmatic method lies in the possibility of interpretation and thus understanding.

By means of our further developed system metamodel, the processual emergence of nex\={u}s and societies through the different regimes, which as we have shown, are completely compatible with object-oriented programming and template meta-programming, can be followed meticulously with the allagmatic method. The method gives guidance for the creation of a complete model with functions, parametrisation, etc. This in turn also gives a better understanding of systems, because specific behaviour and emergent structures can be traced back to meta-theoretical concepts implemented in object-orientated code. The allagmatic method provides an overview because it is a complete framework relating all concepts one to another.

This again is one of Whitehead’s and also Simondon’s important reasons to do metaphysics. The abstract concepts such as society, nexus, structure, operation and so on are supposed to find concrete applications in daily experiences; scientific and non-scientific alike. This is what we have shown in regard to the practice of computer modelling and programming: first these concepts are capable of finding an implementation in object-oriented programming and template meta-programming; second, once the abstract concepts have been successfully implemented, the computations are operating within the context of a broad elaborated allagmatic method of different regimes giving each element its specific place of behaviour and factual arrangement.

\section{Outlook}

Our implementation of adaptation only refers to the emergence of nex\={u}s or societies. It does not show how these nex\={u}s or societies behave or act in regard to an external environment composed of other nex\={u}s and societies. This is not only a further question of implementation but also of interpretation.

First the question arises if only societies survive within a whole computational universe or if also nex\={u}s can survive. This question can be answered with the topic of storage. That is to say, are nex\={u}s simply overwritten and not stored and then dying or are they kept alive by storing them? This can be compared in evolutionary computation with the question whether survival is bound to efficiency in the sense that only the best solutions to a problem are surviving or a randomly selected number of societies is allowed to reproduce. Due to the observed necessity of individuals to be at the same time flexible and specialised in order to survive, it might be useful to design computer models on the basis of a mixture of nex\={u}s and societies as this is already done in genetic algorithms \cite{Holland.1992} and other evolutionary algorithms \cite{Eiben.2015} avoiding too homogeneous populations. 

This in turn would mean to create in Whitehead’s terms so-called structured societies, i.e. societies composed of nex\={u}s and so-called subordinate societies, which are governed and controlled by the overarching structured society. This also means to introduce the concept of hierarchical organisation and reciprocal control: which society is from an operative and structural point of view part of another structured society and so on? Which society is controlling which nex\={u}s and societies? Such description of hierarchical organisation, control, structured societies, and subordinate societies could, for example, then give a deeper look into complex systems and the intricacy of entities.


\end{document}